\documentclass[final]{cvpr}
\pdfoutput=1

\usepackage{times}
\usepackage{epsfig}
\usepackage{graphicx}
\usepackage{amsmath}
\usepackage{amssymb}
\usepackage{booktabs}       %
\usepackage{makecell}
\usepackage{caption} 
\usepackage{subfig}
\usepackage[pagebackref=true,breaklinks=true,colorlinks,bookmarks=false]{hyperref}

\def\cvprPaperID{10709} %
\def\confYear{CVPR 2021}
\begin{document}

\title{Humble Teachers Teach Better Students for Semi-Supervised Object Detection}

\author{Yihe Tang\textsuperscript{~$\dagger$,}\thanks{Work conducted during internship at Amazon Web Services.} \quad Weifeng Chen\textsuperscript{~$\ddagger$} \quad Yijun Luo\textsuperscript{~$\ddagger$} \quad Yuting Zhang\textsuperscript{~$\ddagger$}\vspace{0.2em}\\
\textsuperscript{$\dagger$~}Carnegie Mellon University, \textsuperscript{$\ddagger$~}Amazon Web Services\\
{\tt\small tangacademic@gmail.com \quad \{weifec,yijunl,yutingzh\}@amazon.com
}
\vspace{-0.2em}
}
\maketitle

\thispagestyle{empty}
\begin{abstract}
We propose a semi-supervised approach for contemporary object detectors following the teacher-student dual model framework. Our method
\footnote{Project page: \url{http://yihet.com/humble-teacher}}
is featured with 
1)~the exponential moving averaging strategy to update
the teacher from the student online, 2)~using plenty of region proposals and soft pseudo-labels as the student's training targets, and 
3)~a light-weighted detection-specific data ensemble for the teacher to generate more reliable pseudo-labels. 
Compared to the recent state-of-the-art -- STAC, which uses hard labels on sparsely selected hard pseudo samples, the teacher in our model exposes richer information to the student with soft-labels on many proposals. 
Our model achieves COCO-style AP of $53.04\%$ on \emph{VOC07 val} set, $8.4\%$ better than STAC, when using \emph{VOC12} as unlabeled data. 
On MS-COCO, it outperforms prior work when only a small percentage of data is taken as labeled. 
It also reaches $53.8\%$ AP on \emph{MS-COCO test-dev} with $3.1\%$ gain over the fully supervised ResNet-152 Cascaded R-CNN, by tapping into unlabeled data of a similar size to the labeled data. 
\end{abstract}

\section{Introduction}
We address the problem of semi-supervised object detection in this paper. 
Large curated datasets have driven the recent progress in vision tasks like image classification, 
but data remain scarce for object detection~\cite{girshick2014rich,ren2015faster,liu2016ssd,cai2018cascade,law2018cornernet,redmon2016you}. MS-COCO~\cite{lin2014microsoft}, for example, offers 118,287 annotated images, a relatively small fraction compared to over 14 million labeled images in ILSVRC~\cite{russakovsky2015imagenet}.
Annotation acquisition for detection is also much more costly. 

Much effort has been made to solve the semi-supervised learning problem for image classification, where an object always exists and dominates the image. 
Not all progress for image classification can benefit the detection task significantly as the existence and locations of objects are unknown without bounding box annotations. 
For example, a direct application of classification-based pretraining~\cite{He_2020_moco,chen2020improved} is shown to be not so effective in our experiments (Sec.~\ref{sec:coco_full}).

\begin{figure}[tp!]
  \centering
  \vspace*{0.2em}
  \includegraphics[width=0.75\linewidth]{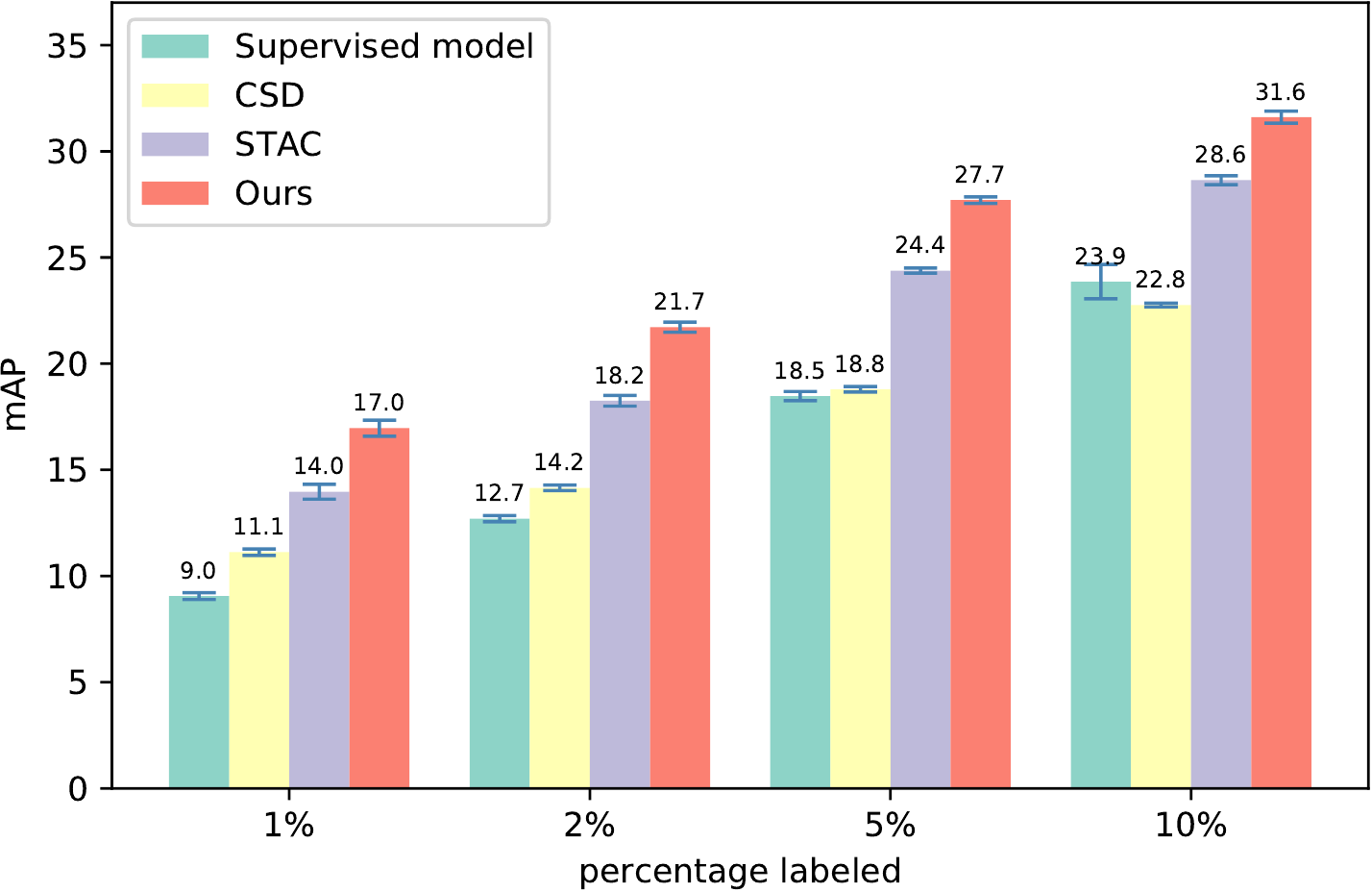}
  \caption{Comparing CSD~\cite{jeong2019consistency}, STAC~\cite{sohn2020simple}, and our approach trained on full \textit{MS-COCO train} 2017 with 1\%, 2\%, 5\%, and 10\% labeled over five runs using the splits in Sec.~\ref{sec:datasets}. Our approach consistently outperforms others.}\label{fig:percentage_COCO}
\end{figure}

In this work, we propose a teacher-student approach called \emph{Humble
Teacher}, which fits modern object detection frameworks better.
The line of work on teacher-student models has many variants, including
self-training~\cite{scudder1965probability,yarowsky1995unsupervised,riloff1996automatically,sohn2020fixmatch,Xie_2020_noisy_student},
the exponential moving average (EMA) based mean teacher~\cite{tarvainen2017mean},
and various ways to obtain pseudo-labels and different views of data
for consistency regularization \cite{zhai2019s4l,laine2016temporal,sajjadi2016regularization,sohn2020fixmatch}
between the teacher and student. Recently, Sohn et al.~\cite{sohn2020simple}
proposed a Self-Training method based on an Augmentation driven Consistency
regularization (STAC) via hard pseudo-labels. It adopted FixMatch~\cite{sohn2020fixmatch},
one of the most successful recent methods for semi-supervised image
classification, directly to the classification head of the Faster
R-CNN~\cite{ren2015faster} detector, yielding improved semi-supervised
detection results. 

Our method further advances the semi-supervised object detection for Faster-R-CNN-like models in a few aspects. 
Unlike self-training with a fixed teacher model, our method updates the teacher model dynamically using EMA updates for object detectors. 
The teacher and student model use asymmetric data augmentation -- stronger augmentations for the student~\cite{xie2019unsupervised,berthelot2019remixmatch,sohn2020fixmatch,sohn2020simple} -- to process different views of the same image~\cite{shorten2019survey}. 
In this framework, the key to our model's strong performance is to use soft pseudo-labels on a reasonable number of region proposals, striking a good balance between covering the entire image and focusing more on learning useful foreground instances. 
It allows the student to distill much richer information from the teacher, compared to sparsely hard-selected high-confident pseudo ground truths in the existing work~\cite{sohn2020simple}. 
The use of soft-labels also keeps the model from over-fitting to the teacher model's potential missing and wrong predictions, which can occur often when using a hard decision threshold. 
In addition, we ensemble the teacher model under a light-weighted detection-specific data augmentation to obtain more reliable pseudo-labels. 
Through our study, we find the wisdom from FixMatch and STAC -- hard pseudo-labels with sample selection -- is not as effective. As our method avoids hard training signals, looks at abundant box instances, seeks for multi-teacher consensus, and uses running average weights as in the mean teacher, we name our method a \emph{Humble Teacher}. 

The humble teacher significantly closes the gap between semi-supervised learning and their fully supervised counterpart on VOC. 
It significantly outperforms the state-of-the-art STAC~\cite{sohn2020simple} on MS-COCO (Fig.~\ref{fig:percentage_COCO}) and VOC by large margins. 
It also improves the ResNet-152 Cascade R-CNN~\cite{cai2018cascade} supervised on \textit{MS-COCO train} significantly with the additional similar-size unlabeled data. 

In summary, we propose the humble teacher for semi-supervised object detection. 
It outperforms the previous state-of-the-art in both low-data and high-data regimes.
Its use of soft-labels are pivotal to enable learning with abundant proposals and also make the EMA and teacher ensemble more effective for detection. 

\section{Related Work}

\subsection{Semi-supervised Learning in Classification}
Significant progress has been made in semi-supervised image classification~\cite{tarvainen2017mean,berthelot2019mixmatch,berthelot2019remixmatch,zhu2002learning,sohn2020fixmatch,laine2016temporal,xie2019unsupervised,iscen2019label}.
One dominant idea in this field is pseudo-labeling~\cite{sohn2020fixmatch, arazo2020pseudo, lee2013pseudo, sohn2020fixmatch,berthelot2019mixmatch,berthelot2019remixmatch,zhai2019s4l,wang2018towards} --- pseudo-labels for unlabeled data are repeatedly generated by a pre-trained model, and the model is then updated by training on a mix of pseudo-labels and human annotated data.
The state-of-the-art FixMatch~\cite{sohn2020fixmatch} retains only the highly confident hard pseudo-labels for training, and adopts different data augmentation strategies for label creation and training.  
Our method draws inspiration from it to use separately augmented inputs for pseudo-labeling and training. Our method is different in that we adopt two separate models --- a student network that learns from pseudo-labels, and a teacher model that annotates pseudo-labels with the aid of a task-specific ensemble. Moreover, we use soft pseudo-labels while~\cite{sohn2020fixmatch} uses hard labels. We additionally update our student and teacher models using two different strategies.

Another popular approach is the consistency regularization~\cite{laine2016temporal,tarvainen2017mean}.
It penalizes the inconsistency between two softmax predictions from different perturbations, such as differently augmented inputs~\cite{laine2016temporal}, prediction and temporal ensemble prediction~\cite{laine2016temporal}.
Our adoption of using soft label is partially inspired by consistency regularization, and extends the soft label idea beyond class probability to also bounding box regression offsets, 
where we keep the predicted offsets of all classes as the soft labels.

A consistency regularization approach, the Mean Teacher~\cite{tarvainen2017mean}, is worth mentioning.
The Mean Teacher adopts a teacher-student approach and the teacher is updated from the student by the exponential moving averaging.
It applies consistency constraints~\cite{tarvainen2017mean} between softmax predictions of the teacher and the student.
Besides being designed for a different detection task, our method looks similar to the Mean Teacher, but there is a critical difference that significantly improves our performance.
Instead of feeding two strongly augmented copies to the teacher and the student for consistency regularization, our teacher sees the original image to make as-accurate-as-possible predictions as pseudo-labels, and our student sees the strongly augmented image to learn more generalizable features.
FixMatch~\cite{sohn2020fixmatch} already demonstrates the big gain of pseudo-labeling compared with the consistency regularization.

\subsection{Semi-supervised Learning in Object Detection}
The pioneering work~\cite{rosenberg2005semi} explores a self-learning approach in object detection based on Mahalanobis metric.
Several works~\cite{gao2019note,hoffman2014lsda,tang2016large} have made progress in utilizing image-level labels to aid semi-supervised object detection.
Adopting ideas similar to those in semi-supervised image classification also leads to progress~\cite{zoph2020rethinking}.
Recently, Sohn et al.~\cite{sohn2020simple} established a new state-of-the-art by combining self-learning and consistency regularization. 
Our work is inspired by it but differs in many ways and attains much better performance. 
First, their approach only has a single network, while we adopt a framework with separate teacher and student networks as in the Mean Teacher~\cite{tarvainen2017mean}.
Second, we generate pseudo-labels from the teacher and train the student simultaneously, while they generate all the pseudo-labels only once and then train on the fixed pseudo-labels.
Third, we use soft labels as the pseudo-labels, while they use hard labels.

\begin{figure*}[htbp]
  \centering
  \includegraphics[width=0.95\linewidth]{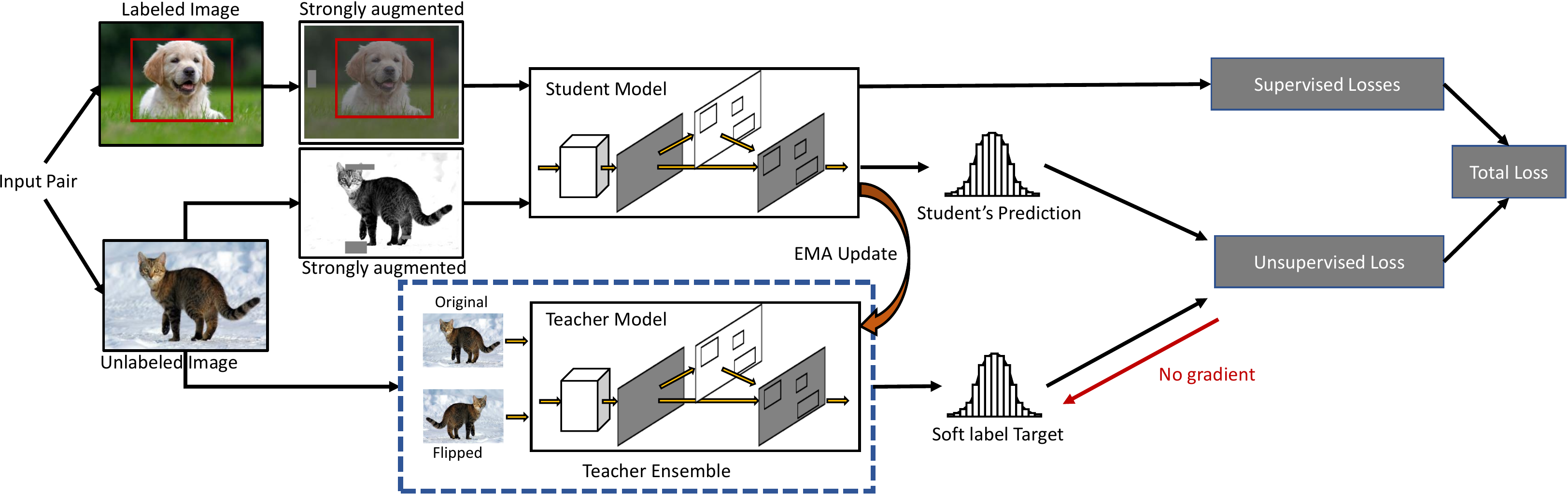}
  \caption{An overview of our Humble Teacher approach. The teacher model produces soft pseudo-labels for the student to learn from, and is updated via exponential moving average (EMA).}\label{fig:arch_overview}
\end{figure*}

Jeong et al. recently proposed CSD~\cite{jeong2019consistency} which horizontally flips an image and enforces its output to be consistent with that from the original image.
CSD inspires our task-specific data ensemble of flipping images for teacher network.
Our idea differs from CSD in the way the flipped images are used: we average the outputs from the original and flipped images to create better pseudo-labels, while CSD uses flipped images to enforce a consistency loss.
Additionally, CSD~\cite{jeong2019consistency} and its follow-up work ISD~\cite{jeong2020interpolation} focus on the grid-sampled boxes in single-stage object detectors, 
while our approach applies to the bounding box proposals in two-stage object detectors such as Faster R-CNN~\cite{ren2015faster}.

\section{Approach}
\label{sec:approach}

\subsection{Overview}
Our approach learns a two-stage object detector from both labeled and unlabeled images.
During training, the framework takes a mixed batch of equal numbers of labeled and unlabeled images as input and feeds them into the supervised branch and the unsupervised branch respectively.
The final loss $L$ is the sum of the supervised loss $L_S$ and the unsupervised loss $L_U$,
\begin{equation}
    L = L_{S} + \frac{n_U}{n_S} \beta L_{U},
\end{equation}
where $n_U, n_S$ are the numbers of unlabeled and labeled images, and $\beta$ is set to $0.5$ by default.

\noindent\textbf{The supervised branch} 
It is a standard supervised two-stage detector like Faster R-CNN~\cite{ren2015faster}.
The regular detection losses are applied ---
the RPN's classification loss $L^\mathrm{rpn}_\mathrm{cls}$ and localization loss $L^\mathrm{rpn}_\mathrm{loc}$, as well as the ROI head's classification loss $L^\mathrm{roi}_\mathrm{cls}$ and localization loss $L^\mathrm{roi}_\mathrm{loc}$.
The total supervised loss is
\begin{equation}
    L_{S} = L^\mathrm{rpn}_\mathrm{cls} + L^\mathrm{rpn}_\mathrm{loc} + L^\mathrm{roi}_\mathrm{cls} + L^\mathrm{roi}_\mathrm{loc}.
\end{equation}

\noindent\textbf{The unsupervised branch} It adopts a teacher-student framework as shown in Fig~\ref{fig:arch_overview}.
The teacher, student, and the supervised network share the same architecture (we use Faster R-CNN~\cite{ren2015faster} in our experiments) and are initialized with the same weights.
The student shares the same weights with the supervised network but not with the teacher.
An unlabeled image is processed independently by both the student and the teacher networks.
The teacher network utilizes a task-specific ensemble to predict a pseudo-label from a weakly augmented version of the image (random flipping).
It only predicts the pseudo-label and does not back-propagate gradients.
The student takes a strongly augmented version of the same image as input to make predictions. 
An unsupervised loss $L_{U}$ is then calculated between the student predictions and the pseudo-labels in RPN and ROI heads.

\noindent\textbf{Augmentation} Augmentation plays an important role in our model.
For training, the image first goes through random flipping and resizing as the weak augmentation.  
The teacher network takes the weakly augmented image as its input (Sec.~\ref{sec:teacher_ensemble}).
Upon the same weakly augmented image, we further randomly change the color, sharpness, contrast, add Gaussian noise and apply cutouts~\cite{devries2017improved}.
We refer to the final image as strongly augmented from the original image. 
Our strong augmentation strategy follows~\cite{sohn2020simple} largely, but we did not use random rotation because bounding boxes will no longer be tightly wrapping around the rotated objects, making the setting undesirably complicated. 
Using strongly augmented images increases the difficulty of the student's task and can encourage it to learn better representations~\cite{shorten2019survey}. 
In contrast, using weak augmentations for the teacher can increase the chance for the teacher to generate correct pseudo-labels. 
Our detailed augmentation method is described in the supplementary material.

\noindent\textbf{Inference Stage} We use the teacher model for inference and produce final object detection results.
No data augmentation is applied to the input image at the inference stage.

\subsection{Soft Labels and Unsupervised Loss}

The unsupervised branch uses soft labels predicted by the teacher model as training targets in the classification and regression tasks.
For the classification task, the soft label target is the predicted distribution of the class probabilities. 
For the bounding box regression task, the soft label target is the offsets of all possible classes when the head is performing class-dependent bounding box regression~\cite{girshick2015fast}. 
We apply unsupervised loss in both the RPN (first stage) and ROI heads (second stage) of our object detector. 
The choice of using soft labels deviates from common practices of using hard labels~\cite{sohn2020simple, sohn2020fixmatch}, where the object categories and offsets are selected when the pseudo-labels are generated.

In the first stage, the unsupervised loss is applied to both the classification objectness and the bounding box regression of the RPN for all anchors $S_A$.
Let $\mathbf{s}^{rpn,i}_\mathrm{cls}$ and $\mathbf{s}^{rpn,i}_\mathrm{reg}$ denote the classification probability and bounding box regression output by the student RPN for the $i$-th proposal, and let $\mathbf{t}^{rpn,i}_\mathrm{cls}$ and $\mathbf{t}^{rpn,i}_\mathrm{reg}$ be those of the teacher RPN. 
Note that the weak augmentations for teacher and student are shared and in sync.
The remaining strong augmentation steps do not impact the image geometry. Consequently, the anchor set is the same for the teacher and the student.
The unsupervised loss for the RPN is defined as
\begin{equation}
    L_{U}^\mathrm{rpn} = \sum_{i\in S_A} D_{KL}(\mathbf{t}^{rpn,i}_\mathrm{cls} \Vert \mathbf{s}^{rpn,i}_\mathrm{cls}) + \| \mathbf{t}^{rpn,i}_\mathrm{reg} - \mathbf{s}^{rpn,i}_\mathrm{reg} \|_2,
    \label{eq:rpn-unsupervised}
\end{equation}
where $D_{KL}$ is the KL divergence. 

In the second stage, the teacher model's RPN generates a set of region proposals, where the standard RPN NMS is applied~\cite{ren2015faster}. 
The teacher model keeps the top-$N$ proposals ranked by the predicted objectness score for the pseudo-label generation.
It is different from the supervised branch, which follows the standard RPN training mode of Faster R-CNN to randomly sample a fix ratio of positive and negative region proposals. 
We set $N=640$ by default and use $S_P$ to denote the set of top-$N$ proposals from the teacher. 
$S_P$ are fed to the ROI heads of both teacher and student. 
The student's RPN proposals are not used in its ROI head training as the teacher's proposals are often of higher quality than those from the student. 
This design also eliminates the need to match proposals between the teacher and student, which could lead to complicated details. 

For each region proposal, the student learns the raw probability and class-dependent regression outputs from the teacher.
Let $\mathbf{s}^{roi,i}_\mathrm{cls}$, $\mathbf{s}^{roi,i}_\mathrm{reg}$, $\mathbf{t}^{roi,i}_\mathrm{cls}$, $\mathbf{t}^{roi,i}_\mathrm{reg}$ denote the classification probabilities and all-class bounding box regression outputs by the student and teacher ROI head for the $i$-th proposal respectively, our final ROI consistency loss is
\begin{equation}
    L_{U}^\mathrm{roi} = \Sigma_{i\in S_P} D_{KL}(\mathbf{t}^{roi,i}_\mathrm{cls} \Vert \mathbf{s}^{roi,i}_\mathrm{cls}) + \| \mathbf{t}^{roi,i}_\mathrm{reg} - \mathbf{s}^{roi,i}_\mathrm{reg} \|_2.
\end{equation}
The final unsupervised loss $L_{U}$ is the sum of $L_{U}^\mathrm{roi}$ and $L_{U}^\mathrm{rpn}$.

The use of all top-$N$ regions proposals results in abundant box instances for pseudo-labels. 
They are likely to cover the actual objects, boxes moderately overlapped with objects, and background regions, leading to a more comprehensive representation of the detection score distribution over the entire image. 
These benefits are unattainable when using hard labels. Many regions are neither strictly foreground nor background, and the hard labels cannot represent such intermediate states. 
The hard label setting, such as in~\cite{sohn2020simple}, naturally needs a sample selection process like NMS and score-based thresholding to get definite pseudo ground truths.

\subsection{Exponential Moving Average for the Teacher Model Update}

The teacher model weights $W_\mathrm{teacher}$ are updated from the student model weights $W_\mathrm{student}$ by exponential moving average (EMA)~\cite{tarvainen2017mean}.
At each iteration, we have
\begin{equation}
    W_\mathrm{teacher} = \alpha W_\mathrm{teacher} + (1-\alpha)W_\mathrm{student},
\end{equation}
where we set $\alpha=0.999$.
Therefore, the teacher only slightly updates itself from the student each time.
The gradually updated teacher is more resilient to the sudden weight turbulence of the student due to a wrong label prediction of the teacher model ---  even if the student is fed with a wrong label, its influence on the teacher model is mitigated by the exponential moving average.
Besides resiliency to occasional wrong pseudo-labels, EMA is also known to lead to better generalization~\cite{izmailov2018averaging}.

It is worth noting that we follow Faster R-CNN~\cite{ren2015faster} to fix the running mean and variance of the BatchNorm layers in the training.

\begin{table*}%
  \centering
  \begin{tabular}{lllcc}
    \toprule
    Model & Labeled Dataset & Unlabeled Dataset & AP50 & AP  \\
    \midrule
    Supervised model & VOC07 & N/A & 76.3 & 42.60 \\
    Supervised model & VOC07 + VOC12 & N/A & 82.17 & 54.29\\
    \midrule
    CSD$^\ddag$ & VOC07 & VOC12 & 76.76 & 42.71 \\
    STAC~\cite{sohn2020simple} & VOC07 & VOC12 & 77.45 & 44.64\\
    \textbf{Humble teacher (ours)} & VOC07 & VOC12 & \textbf{80.94} & \textbf{53.04}\\
    \midrule
    CSD$^\ddag$ & VOC07 & VOC12 + MS-COCO20 (2017) & 77.10 & 43.62\\
    STAC~\cite{sohn2020simple} & VOC07 & VOC12 + MS-COCO20 (2017) & 79.08 & 46.01\\
    \textbf{Humble teacher (ours)} & VOC07 & VOC12 + MS-COCO20 (2017)& \textbf{81.29} & \textbf{54.41}\\
    \bottomrule
  \end{tabular}
  \caption{Results on Pascal VOC, evaluated on the \emph{VOC07 test} set. Our model consistently outperforms others in all experiment setups. CSD$^\ddag$~is our ResNet-50-based re-implementation, which achieves better performance than the original CSD~\cite{jeong2019consistency}.}
  \label{tab:pascal_voc}
\end{table*}

\subsection{Teacher Ensemble with Horizontal Flipping}
\label{sec:teacher_ensemble}
We ensemble the teacher model by taking as input both the image and its horizontally flipped version.
The underlying intuition is that object classes should remain the same when the image is flipped, and the average prediction from both the original and the flipped copy can be more accurate than the prediction from a single image. Our design is inspired by prior research on ensemble methods~\cite{rokach2010ensemble,radosavovic2018data,zou2019learning}, and by human pose estimation literature in which combining predictions from the original and the flipped image has lead to better pose estimation~\cite{newell2016stacked,sun2018integral,chen2018cascaded}.
Experiments in Sec.~\ref{sec:exp:structual_ensemble} show that our teacher ensemble leads to superior semi-supervised object detection performance.

More specifically, let $f_B$ be the backbone feature of the original image, $\hat{f_B}$ be the backbone feature of the flipped image, and $P$ be the set of proposals detected by RPN on the original image.
We do not use RPN to propose regions for the flipped image but instead flip the proposal coordinates in $P$ horizontally to obtain $\hat{P}$ as the proposals for the flipped image. 
Then, for the ROI head, its softmax class probability output $P_\mathrm{cls}$ and regression offset output $\sigma_\mathrm{reg}$ from the ensemble are:
\begin{align}
    f &= \mathrm{ROIAlign}(f_B, P),\\
    \hat{f} &= \mathrm{ROIAlign}(\hat{f_B}, \hat{P}),\\
    P_\mathrm{cls} &= 0.5(C(f) + C(\hat{f})),\\
    \sigma_\mathrm{reg} &= 0.5(R(f) + T(R(\hat{f}))).
\end{align}
Note that $C$ is the classification head including softmax at the end, and $R$ is the regression head.
$T$ is the transformation that flips the $x$ axis of all bounding boxes.
We apply this ensemble mechanism only to create pseudo-labels in the ROI heads but not RPN heads, because the corresponding anchors in a flipped pair of images may not be symmetric in the RPN head.

\begin{table*}%
  \centering
  \begin{tabular}{lrrrr}
    \toprule
    Percentage labeled & 1\% & 2\% & 5\% & 10\% \\
    \midrule
    Supervised model & 9.05$\pm$0.16 & 12.70$\pm$0.15 & 18.47$\pm$0.22 & 23.86$\pm$0.81 \\
    CSD$^\ddag$ & 11.12$\pm$0.15 ($+$2.07) & 14.15$\pm$0.13 ($+$1.45) & 18.79$\pm$0.13 ($+$0.32) & 22.76$\pm$0.09 ($-$1.10)\\
    STAC~\cite{sohn2020simple} & 13.97$\pm$0.35 ($+$4.92) & 18.25$\pm$0.25 ($+$5.55) & 24.38$\pm$0.12 ($+$5.91) & 28.64$\pm$0.21 ($+$4.78)\\
    \textbf{Humble teacher (ours)} & \textbf{16.96$\pm$0.38 ($+$7.91)} & \textbf{21.72$\pm$0.24 ($+$9.02)} & \textbf{27.70$\pm$0.15 ($+$9.23)} & \textbf{31.61$\pm$0.28 ($+$7.74)}\\
    \bottomrule
  \end{tabular}
  \caption{The mAP (50:95) results on \textit{MS-COCO val 2017} by models trained on different percentage of labeled \textit{MS-COCO train 2017}. All models are with the ResNet-50 backbone. CSD$^\ddag$~is our re-implementation with better performance. Our method consistently outperforms others.}
  \label{tab:MS-COCO_percentage}
\end{table*}

\section{Experiments}

\subsection{Dataset and Evaluation}
\label{sec:datasets}

We evaluate our approach on two detection datasets: Pascal VOC~\cite{everingham2010pascal} and MS-COCO \cite{lin2014microsoft}.
For Pascal VOC, we evaluate the performance on the \textit{VOC07 test}.
During training, we first use \textit{VOC07 trainval} as the labeled dataset and \textit{VOC12 trainval} as the unlabeled dataset.
\textit{VOC07 trainval} and \textit{VOC12 trainval} have 5,011 and 11,540 images respectively, resulting in a roughly 1:2 labeled to unlabeled ratio.
Following the practice in \cite{jeong2019consistency, sohn2020simple}, besides \textit{VOC12 trainval}, we also bring \textit{MS-COCO20}~\cite{jeong2019consistency, sohn2020simple} in as additional unlabeled data.
\textit{MS-COCO20} filters out the MS-COCO images that contain objects whose classes are not included in the 20 Pascal VOC classes.
We conduct additional experiments using both the \textit{VOC12 trainval} and \textit{MS-COCO20 train} as unlabeled data, totaling 129,827 unlabeled images, leading to a 1:26 labeled to unlabeled ratio.

For MS-COCO, we use version 2017 in all experiments. We report the results on the \textit{MS-COCO val} dataset.
For training, we follow~\cite{sohn2020simple} to split \textit{MS-COCO train} into the labeled and the unlabeled datasets.
We set up four labeling percentages: 1\%, 2\%, 5\%, and 10\% as in \cite{sohn2020simple}, and the remaining images are used as unlabeled data.
For each percentage, we randomly sample five different splits using the provided code from~\cite{sohn2020simple}. 
The same splits are used throughout our experiments and ablation studies.
In addition, we also set up an experiment using the entire \textit{MS-COCO train} as labeled dataset, and \textit{MS-COCO unlabeled} as unlabeled dataset.
\textit{MS-COCO train} has a total of 118,287 images and \textit{MS-COCO unlabeled} has 123,403 in total, leading to a roughly 1:1 labeled to unlabeled ratio.
We run this experiment to demonstrate that our approach is able to further improve upon a model trained on a large labeled dataset like MS-COCO.

\subsection{Model Configurations}

We use Faster R-CNN with ResNet-50 backbone and FPN as our default base model.
We re-implement CSD with ResNet-50 backbone for fair comparison, and it achieves better performance than the original model in \cite{jeong2019consistency}.
We also evaluate our method on a larger base model Cascade R-CNN with ResNet-151 backbone and FPN~\cite{cai2018cascade,lin2017feature}.
When training on Cascade R-CNN, we apply our unsupervised loss on the ROI head at each stage .

Before training on unlabeled data, the model first goes through a \emph{burn-in} stage, i.e. pre-training the detection network on the labeled data following standard training protocols~\cite{ren2015faster}. This model is the base supervised model, and its weights are copied into the student and the teacher networks to initiate the semi-supervised training.

\subsection{Results on Pascal VOC}

We benchmark our method on PASCAL VOC under two experiment setups --- \textbf{(a)} \textit{VOC07} as labeled set and \textit{VOC12} as unlabeled set, and \textbf{(b)} the same as (a) but with \textit{MS-COCO20} as additional unlabeled data. We also report the performance of the same model trained fully supervised on \textit{VOC07} and \textit{VOC07+VOC12}. Tab.~\ref{tab:pascal_voc} compares our results with the best existing methods under AP50 and MS-COCO style AP metrics.

Our approach consistently outperforms the best existing results by a large margin in all setups. It outperforms the state-of-the-art STAC~\cite{sohn2020simple} by 8.4\% and 8.4\% in AP respectively in setup (a) and (b). Notably, our method trained on the labeled \textit{VOC07} and the unlabeled \textit{VOC12} significantly outperforms the based model fully supervised on \textit{VOC07} alone, and with the additional unlabeled \textit{MS-COCO20} it further improves performance. Our best performing model is narrowing the gap from 9.65\% to 1.25\% in COCO style mAP between the model fully supervised on \textit{VOC07+VOC12} and the model trained on labeled \textit{VOC07} and unlabeled \textit{VOC12}. These results suggest that our method is particularly effective in improving model performance with cheap unlabeled data.

Moreover, our model outperforms CSD and STAC more on the 0.5:0.95 AP than on AP50 regarding both absolute gain and relative error reduction.
It indicates that the humble teacher could localize objects more accurately. 
This may be attributed to the use of soft labels over the full set of region proposals, which leads to more guidance for the student model to learn on image regions without definite labels even given the ground truth annotations. 
Such guidance has been shown to be helpful for localization~\cite{zhang2015lmdis}.

\subsection{Results on MS-COCO}
\label{exp:res_on_coco}

\subsubsection{MS-COCO of Different Labeled Percentages}
\label{sec:coco_percentage}
We first investigate if the proposed humble teacher improves performance under a low data regime. We follow the setup of STAC~\cite{sohn2020simple} and report the performance when four percentages of labeled \textit{MS-COCO train} is provided: 1\%, 2\%, 5\% and 10\%, while the remaining images are used as unlabeled data. Comparison with the best existing approaches on \textit{MS-COCO val} in terms of mAP (50:95) is shown in Tab.~\ref{tab:MS-COCO_percentage}.
Our method consistently outperforms the best existing approach over all four labeled percentages.
Notably, unlike CSD, the amount of improvement does not diminish, and the improvement is consistent though the percentage of labeled data increases.

\begin{table}%
  \centering
  \resizebox{\columnwidth}{!}{
  \begin{tabular}{lr}
    \toprule
    Model (Faster R-CNN with Resnet-50) &  AP\\
    \midrule
    Base supervised model & 37.63 \\
    MOCOv2 + MS-COCO Unlabeled~\cite{chen2020improved}      & 35.29 \\
    MOCOv2 + ImageNet-1M~\cite{chen2020improved}  & 40.80 \\
    MOCOv2 + Instagram-1B~\cite{chen2020improved} & 41.10 \\
    
    Proposal learning~\cite{tang2021proposal} & 38.4 \\
    CSD$^\ddag$ & 38.52($+$0.89) \\
    STAC~\cite{sohn2020simple} & 39.21($+$1.58) \\
    \textbf{Humble teacher (ours)} & \textbf{42.37($+$4.74)}\\
    \toprule
    Model (Cascade R-CNN with ResNet-152)&  AP\\
    \midrule
    Base supervised model & 50.23 \\
    \textbf{Humble teacher (ours)} & \textbf{53.38 ($+$3.15)}\\
    \bottomrule
  \end{tabular}
  }
  \caption{The mAP (50:95) results on \textbf{\textit{MS-COCO val 2017}} by models trained on \textit{MS-COCO train 2017} + \textit{MS-COCO unlabeled}. CSD$^\ddag$~is with a ResNet-50 backbone.}
  \label{tab:MS-COCO_unlabeled_exps}
\end{table}

\begin{table}%
 
  \label{tab:MS-COCO_unlabeled_exps_2}
  \centering
  \resizebox{\columnwidth}{!}{
  \begin{tabular}{lr}
    \toprule
    Model (Cascade R-CNN with ResNet-152)&  AP\\
    \midrule
    Base supervised model & 50.7 \\
    \textbf{Humble teacher (ours)} & \textbf{53.8 ($+$3.1)}\\
    \bottomrule
  \end{tabular}
  }
  \caption{The mAP (50:95) results on \textbf{\textit{MS-COCO test-dev 2017}} by models trained on \textit{MS-COCO train 2017} + \textit{MS-COCO unlabeled}.}
  \label{tab:MS-COC_test_dev}
\end{table}

\subsubsection{MS-COCO Train + MS-COCO Unlabeled} 
\label{sec:coco_full}

Next, we investigate if the proposed semi-supervised learning strategy improves upon an object detector fully supervised on the entire \textit{MS-COCO train}. We use the \textit{MS-COCO unlabeled}~\cite{lin2014microsoft}, a set of 123,403 unlabeled images differed from those in \textit{MS-COCO train}. We experiment with two setups, one is with Faster R-CNN~\cite{ren2015faster} and another with Cascade R-CNN~\cite{cai2018cascade}. The results are evaluated on \textit{MS-COCO val}. In the Faster R-CNN case, the baseline model supervised on the full \textit{MS-COCO train} achieves 37.63\% AP. Our method achieves a 4.74\% improvement in AP over the baseline (Tab.~\ref{tab:MS-COCO_unlabeled_exps}), and significantly outperforms other self-supervised methods such as Proposal Learning~\cite{tang2021proposal}, CSD~\cite{jeong2019consistency} and STAC~\cite{sohn2020simple}. In the Cascade R-CNN case, our method achieves a 3.15\% improvement in AP over the high-performing fully supervised baseline (Tab.~\ref{tab:MS-COCO_unlabeled_exps}). 
Further evaluation on the \textit{MS-COCO test-dev} shows a 3.1\% AP improvement over the supervised Cascade R-CNN (Tab.~\ref{tab:MS-COC_test_dev}).
These results suggest that our method has the potential to directly apply to any object detectors and improve their performance by combining both labeled and unlabeled data. 

We also compare against supervised finetuning with pretrained MOCOv2~\cite{chen2020improved}, a state-of-the-art contrastive learning method for image classification pretraining. The goal is to show that a simple application of contrastive learning~\cite{chen2020improved,He_2020_moco,chen2020simple} does not work as well as our method in improving object detection from unlabeled data. 
More specifically, we follow the MOCOv2 setup to pre-train the ResNet-50 backbone in Faster R-CNN on each of the three unlabeled dataests: (1) \textit{MS-COCO unlabeled}, (2) ImageNet-1M~\cite{deng2009imagenet} and (3) Instagram-1B~\cite{mahajan2018exploring}. The pre-trained backbones are then copied to Faster R-CNN, which is further trained on \textit{MS-COCO train} to perform object detection. Results in Tab.~\ref{tab:MS-COCO_unlabeled_exps} suggest that object detection performance improves as the size of the unlabeled data increases. However, even the best-performing one (MOCOv2 pre-trained on Instagram-1B) still underperforms our method, although it uses 7,600 times more unlabeled data than our method. 

\section{Ablation Study}
\subsection{Number of Proposals for Unsupervised Loss}
\label{sec:num-proposals}
We first study how the number of region proposals fed into ROI head in unsupervised learning affects the performance.
As shown in Fig.~\ref{fig:number_of_proposals}, we experiment with different numbers of proposals up to 6000 given the GPU memory limit.
We found that using too few region proposals hurts performance, possibly because of a poor coverage of objects and useful context. 
Having too many region proposals may include too many background samples, distracting the unsupervised learning from the important foreground regions~\cite{jeong2019consistency}.
Given the large performance drop when the proposals are too few or too many, we believe that using a balanced number of proposals with soft labels is the key to the superior performance of our method. 

\subsection{Update Rules}

This section studies the benefits of our EMA update at every iteration.   
The teacher model is updated from the student model. We study three rules with different update frequencies: 
(1) EMA update at every iteration, (2) copy weights from student to teacher every 10K iterations, and (3) no update at all, i.e. keeping the teacher model fixed throughout the training.
We still use Faster R-CNN with ResNet-50 for all the rules and trained on 10\% labeled \textit{MS-COCO train 2017}. Tab.~\ref{tab:MS-COCO_update_rules} reports the mean and standard deviations over five runs using the same five splits described in Sec.~\ref{sec:datasets}.

\begin{table}[b]%
  \centering
  \resizebox{\columnwidth}{!}{
  \begin{tabular}{lr}
    \toprule
    Model &  AP\\
    \midrule
    No update & 27.26$\pm$0.21\\
    Copy weights from student to teacher every 10K iters & 28.61$\pm$0.18\\
    \textbf{EMA update at every iter} & \textbf{31.61$\pm$0.28} \\
    \bottomrule
  \end{tabular}
  }
  \caption{Comparison between different update rules on \textit{MS-COCO train 2017} with 10\% data labeled. The mean and standard deviation over five data splits are reported (the same five splits of \textit{MS-COCO train 2017} as in Sec.~\ref{sec:datasets}).}
  \label{tab:MS-COCO_update_rules}
\end{table}

\begin{figure*}
    \begin{centering}
    \subfloat[Comparison between models with different number of region proposals used in unsupervised loss. The student-teacher framework is jointly trained on the 10\% labeled and 90\% unlabeled \textit{MS-COCO train 2017} split.]{\begin{centering}
        \includegraphics[width=0.305\textwidth]{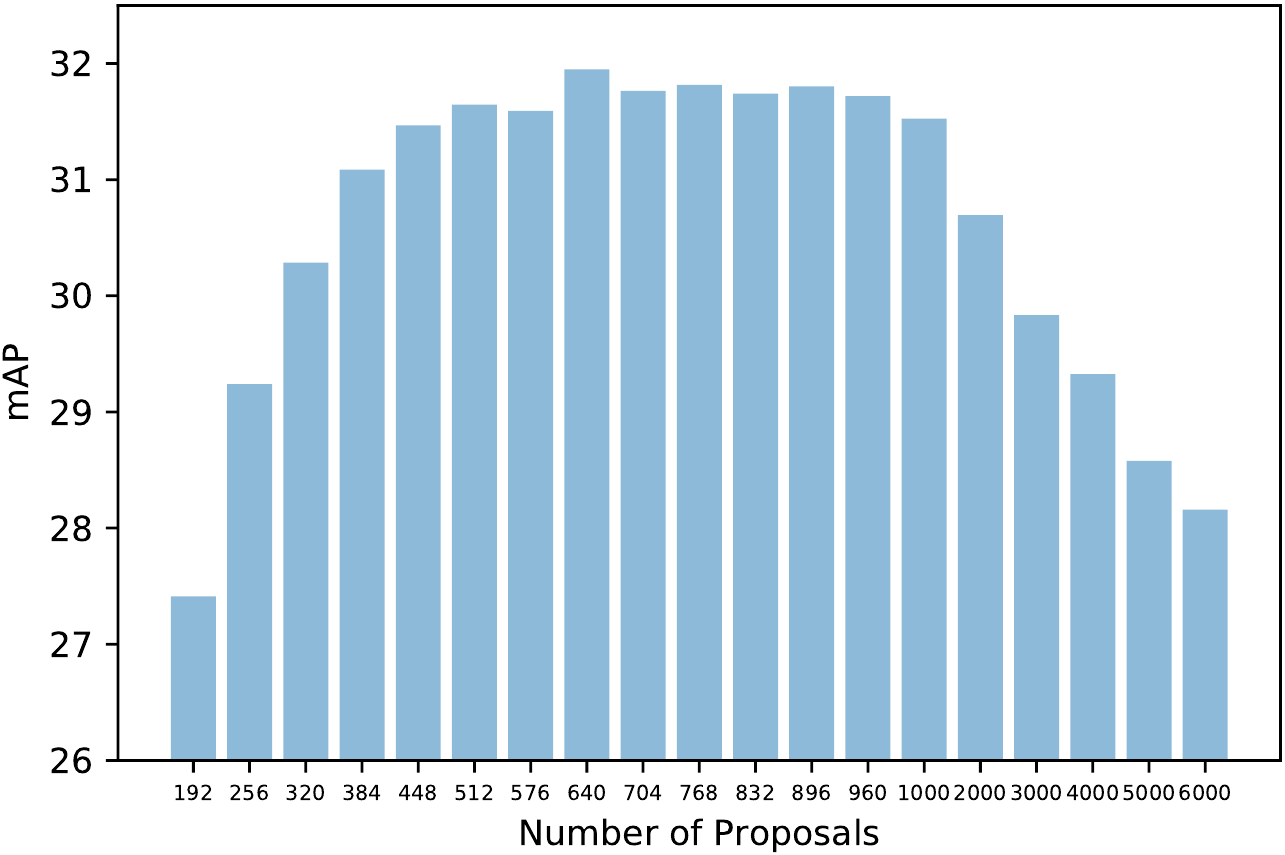}
        \label{fig:number_of_proposals}
    \par\end{centering}
    }\hspace*{\fill}\subfloat[Comparison between teacher and student performance on the 10\% labeled \textit{MS-COCO train 2017} setup. The student-teacher framework is jointly trained on the 10\% labeled and 90\% unlabeled \textit{MS-COCO train 2017} split.]{\begin{centering}
        \includegraphics[width=0.305\textwidth]{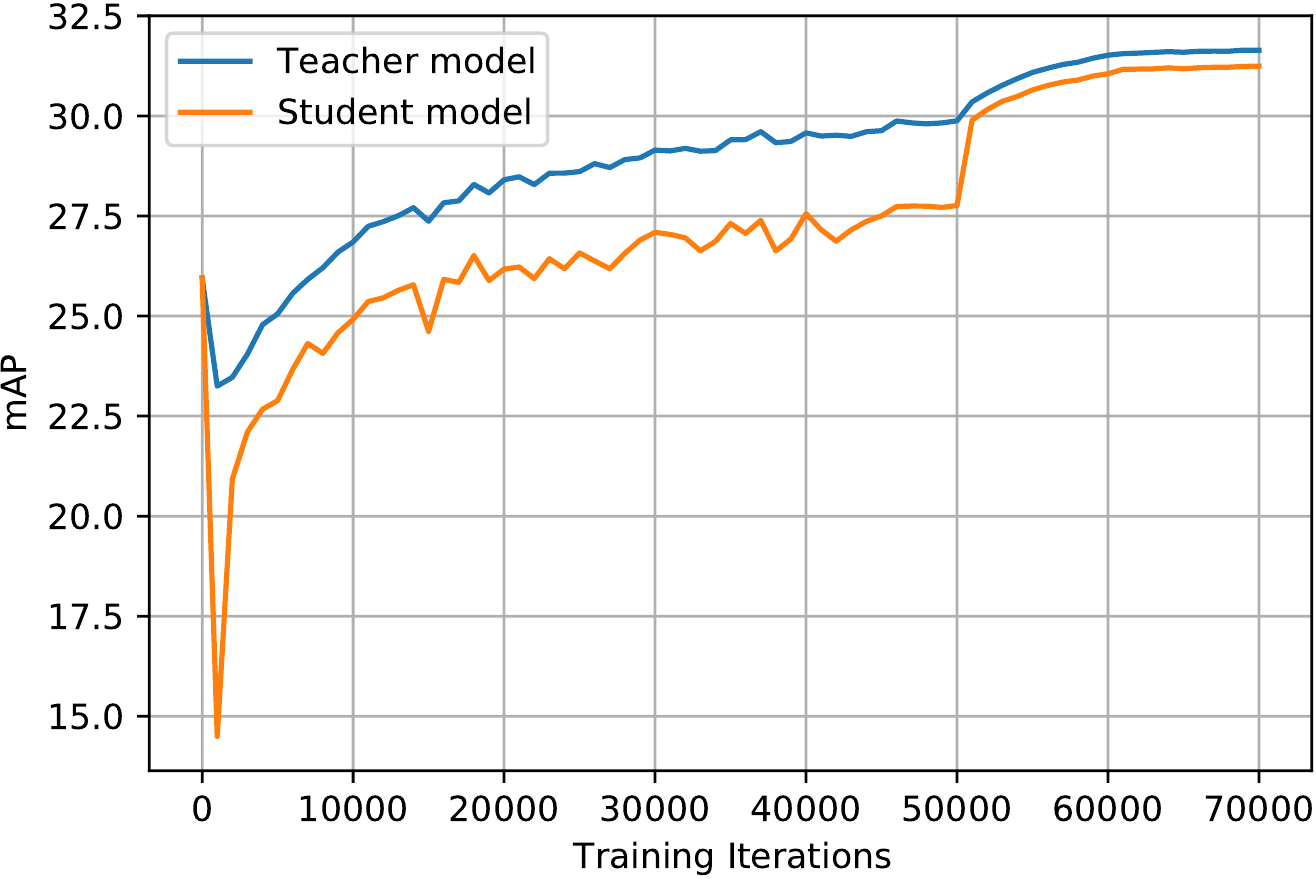}
        \label{fig:gt_ap_student_vs_teacher}
    \par\end{centering}
    }\hspace*{\fill}\subfloat[Teacher models' performance on unlabeled data. Both models are trained on 10\% labeled \textit{MS-COCO train 2017} with the remaining 90\% as unlabeled.]{\begin{centering}
    \includegraphics[width=0.305\textwidth]{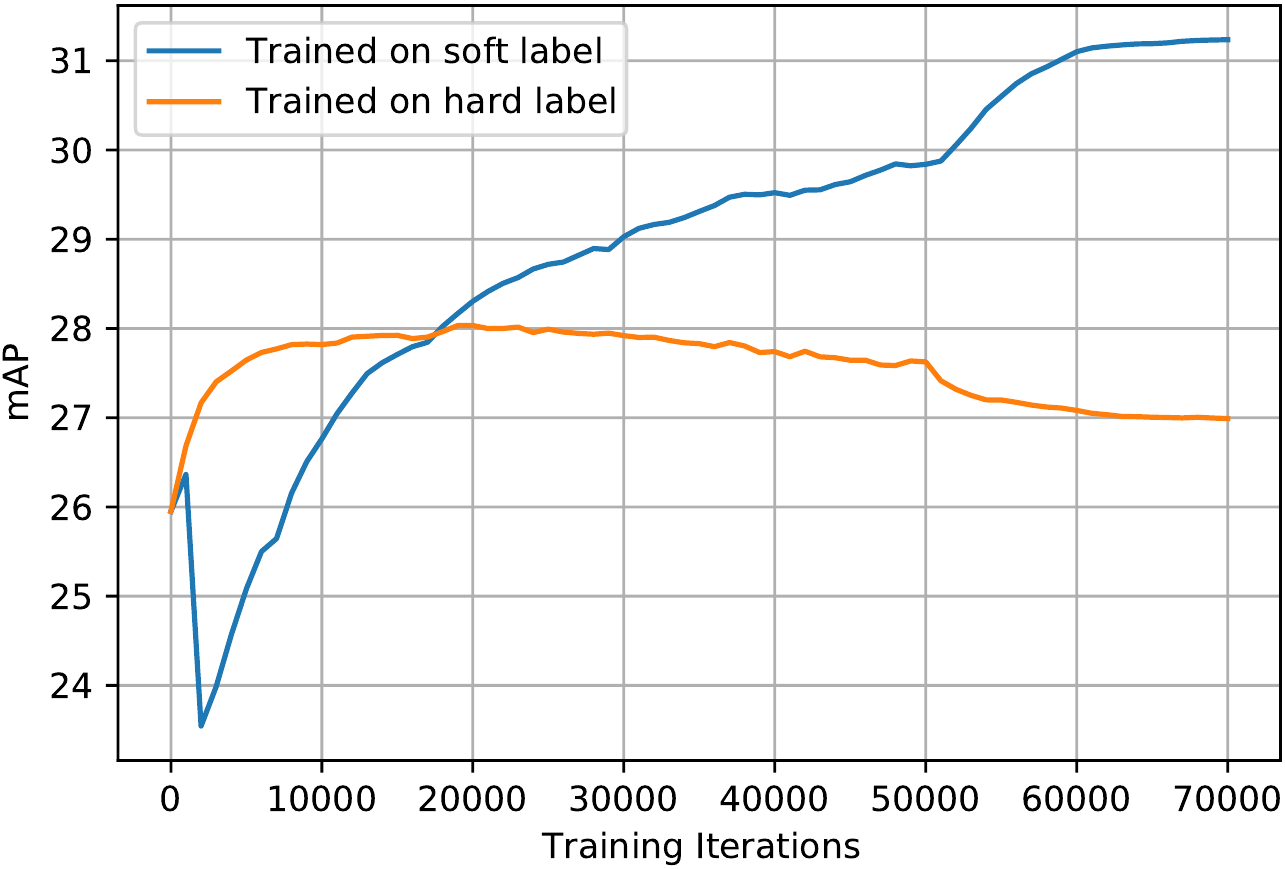}
        \label{fig:gt_ap_soft_vs_hard}
    \par\end{centering}
    }
    \par\end{centering}
    \caption{Ablation study on hyperparameters and hard/soft labels.}

\end{figure*}

Updating every 10K iterations outperforms no update at all. It suggests that keeping the teacher model up to date than using a fixed teacher is beneficial to model performance. EMA update at every iteration leads to even bigger performance gain. The results suggest that EMA updates are crucial for our student-teacher model to work well. 
One possible explanation is that the negative effect of incorrect pseudo-labels is mitigated by EMA update at every iteration, since the weight updates from one example batch are being averaged over time and sample batches.

The success of EMA is based on the assumption that EMA-updated teacher produces more accurate predictions than the student. To validate this assumption, we compare the object detection results on the 10\% labeled \textit{MS-COCO train 2017} setup using the student and the EMA-updated teacher model. 
Fig.~\ref{fig:gt_ap_student_vs_teacher} shows that the EMA-updated teacher is better than the student and therefore explains the success of our student-teacher paradigm.

\subsection{Soft Labels versus Hard Labels} 
\label{sec:abl:soft_label}

Next, we turn to the comparison between soft labels and hard labels in our semi-supervised framework. 
We use the same Faster R-CNN with ResNet-50 setup as before, and train on the 10\% labeled \textit{MS-COCO train 2017}, 
using the same EMA update and teacher-student framework. 
We then compare a version that trains on soft labels and another that trains on hard labels.
Note that the hard labels are generated by thresholding on the prediction confidence. 
We experiment with a range of thresholds and select 0.7 which leads to the best performance.
Moreover, given it is unclear how to combine the hard label from an original image and its flipped version, 
we exclude the task specific data ensemble from both experiments for fairness of comparison.
Tab.~\ref{tab:softlabel_1} reports the results. 
Contrary to the findings in semi-supervised image classification~\cite{sohn2020fixmatch}, using soft labels help us achieve much better performance than hard labels in semi-supervised object detection, clearly demonstrating the critical role of soft labels plays in our method.

\begin{table}[b]%
  \centering
  \begin{tabular}{lr}
    \toprule
    Model &  AP\\
    \midrule
    With hard label & 27.97$\pm$0.13\\
    \textbf{With soft label} & \textbf{30.97$\pm$0.16}\\
    \bottomrule
  \end{tabular}
   \caption{Comparison between training on soft label and hard label when 10\% labeled \textit{MS-COCO train 2017} is provided. The mean and standard deviation over five data splits are reported (the same five splits of \textit{MS-COCO train 2017} described in Sec.~\ref{sec:datasets}).}
   \label{tab:softlabel_1}
\end{table}

One possible explanation to the better performance due to the soft label is its strength to handle the highly imbalanced class distribution in object detection. 
This imbalanced issue is reflected in two aspects. 
First, background class dominates foreground classes during region proposal~\cite{ren2015faster}. 
Second, foreground classes are not evenly distributed during ROI classification, as evident in the case of MS-COCO \cite{lin2014microsoft}. 
Using hard labels in such an imbalanced setup has the risk of pushing the probability of being dominant classes to 1 and the probability of being minority classes to 0, resulting in significant confirmation bias~\cite{confirmation_bias}. In contrast, soft labels carry richer information and retain the probability of being any possible classes, and suffer from less confirmation bias.

To validate this hypothesis, we experiment with the 10\% labeled \textit{MS-COCO train 2017} setup, and run object detection every 1,000 iterations using the teacher model on the remaining 90\% unlabeled images and evaluate the detection mAP. Fig.~\ref{fig:gt_ap_soft_vs_hard} reports the mAP as training proceeds. We see that training on soft labels yields much higher mAP, and the mAP keeps increasing as the training goes on, while training on hard labels yields diminishing mAP. These results indicate that soft-label-trained teachers produce pseudo-labels that suffer from less confirmation bias. 

\subsection{Teacher Ensemble}
\label{sec:exp:structual_ensemble}

We study the effectiveness of Teacher Ensemble. 
FixMatch~\cite{sohn2020fixmatch} and ReMixMatch~\cite{berthelot2019remixmatch} claim a data ensemble of random augmentations may hurt the teacher model performance, and is worse than weak augmentations (resizing and randomly flipping) applied to the inputs of the teacher.
We find this partially true, and show that our teacher ensemble improves performance in semi-supervised object detection.

Our experiment is based on the same Faster R-CNN with ResNet-50 trained on 10\% labeled \textit{MS-COCO train 2017}. We compare three setups: (1) without ensemble, (2) with a random augmented ensemble on teacher model, and (3) with task-specific data ensemble on teacher model. Tab.~\ref{tab:structural_ensemble} reports the results.

\begin{table}[htbp]
  
  \centering
  \begin{tabular}{lr}
    \toprule
    Model &  AP\\
    \midrule
    No ensemble & 30.97$\pm$0.16 \\
    Random augmented ensemble & 30.79$\pm$0.31\\
    \textbf{Task-specific ensemble } & \textbf{31.61$\pm$0.28}\\
    \bottomrule
  \end{tabular}
   \caption{Effects of using different ensemble strategies on the teacher model on \textit{MS-COCO train 2017} with 10\% data labeled. The mean and standard deviation over five data splits are reported (the same five splits of \textit{MS-COCO train 2017} described in Sec.~\ref{sec:datasets}).}
   \label{tab:structural_ensemble}
\end{table}

Consistent with the findings in FixMatch~\cite{sohn2020fixmatch}, the random augmentation ensemble indeed hurts performance.
Nonetheless, with our task-specific data ensemble (ensembling a pair of flipped and original images), the performance improves by 0.64\% AP,
suggesting that a carefully constructed ensemble is advantageous to the overall performance of our semi-supervised object detection method.

\section{Conclusions}
We developed a semi-supervised object detection algorithm, ``Humble Teacher'' that obtained state-of-the-art performance on multiple benchmarks.
We demonstrated the effectiveness of our teacher-student model design and showed the importance of iteration-wise EMA teacher update.
We found that soft label coupled with a balanced number of teacher's region proposals is the key toward superior performance. 
We also found that a carefully constructed data ensemble for the teacher improves the overall performance.

\noindent\textbf{Acknowledgements}
We want to thank Luis Goncalves, Zhaowei Cai, Qi Dong, Aruni RoyChowdhury, R. Manmatha, Zhuowen Tu, and Vijay Mahadevan for insightful discussions.

{\small
\bibliographystyle{ieee_fullname}
\bibliography{egbib}

\begin{thebibliography}{10}\itemsep=-1pt

\bibitem{arazo2020pseudo}
Eric Arazo, Diego Ortego, Paul Albert, Noel~E O’Connor, and Kevin McGuinness.
\newblock Pseudo-labeling and confirmation bias in deep semi-supervised
  learning.
\newblock In {\em 2020 International Joint Conference on Neural Networks},
  pages 1--8. IEEE, 2020.

\bibitem{confirmation_bias}
Eric Arazo, Diego Ortego, Paul Albert, Noel~E O’Connor, and Kevin McGuinness.
\newblock Pseudo-labeling and confirmation bias in deep semi-supervised
  learning.
\newblock In {\em 2020 International Joint Conference on Neural Networks},
  pages 1--8, 2020.

\bibitem{berthelot2019remixmatch}
David Berthelot, Nicholas Carlini, Ekin~D Cubuk, Alex Kurakin, Kihyuk Sohn, Han
  Zhang, and Colin Raffel.
\newblock {ReMixMatch}: Semi-supervised learning with distribution alignment
  and augmentation anchoring.
\newblock {\em arXiv preprint arXiv:1911.09785}, 2019.

\bibitem{berthelot2019mixmatch}
David Berthelot, Nicholas Carlini, Ian Goodfellow, Nicolas Papernot, Avital
  Oliver, and Colin~A Raffel.
\newblock {MixMatch}: A holistic approach to semi-supervised learning.
\newblock In {\em Advances in Neural Information Processing Systems}, pages
  5049--5059, 2019.

\bibitem{cai2018cascade}
Zhaowei Cai and Nuno Vasconcelos.
\newblock Cascade {R-CNN}: Delving into high quality object detection.
\newblock In {\em Proceedings of the IEEE Conference on Computer Vision and
  Pattern Recognition}, pages 6154--6162, 2018.

\bibitem{chen2020simple}
Ting Chen, Simon Kornblith, Mohammad Norouzi, and Geoffrey Hinton.
\newblock A simple framework for contrastive learning of visual
  representations.
\newblock {\em arXiv preprint arXiv:2002.05709}, 2020.

\bibitem{chen2020improved}
Xinlei Chen, Haoqi Fan, Ross Girshick, and Kaiming He.
\newblock Improved baselines with momentum contrastive learning.
\newblock {\em arXiv preprint arXiv:2003.04297}, 2020.

\bibitem{chen2018cascaded}
Yilun Chen, Zhicheng Wang, Yuxiang Peng, Zhiqiang Zhang, Gang Yu, and Jian Sun.
\newblock Cascaded pyramid network for multi-person pose estimation.
\newblock In {\em Proceedings of the IEEE Conference on Computer Vision and
  Pattern Recognition}, pages 7103--7112, 2018.

\bibitem{deng2009imagenet}
Jia Deng, Wei Dong, Richard Socher, Li-Jia Li, Kai Li, and Li Fei-Fei.
\newblock {ImageNet}: A large-scale hierarchical image database.
\newblock In {\em 2009 IEEE Conference on Computer Vision and Pattern
  Recognition}, pages 248--255. Ieee, 2009.

\bibitem{devries2017improved}
Terrance DeVries and Graham~W Taylor.
\newblock Improved regularization of convolutional neural networks with cutout.
\newblock {\em arXiv preprint arXiv:1708.04552}, 2017.

\bibitem{everingham2010pascal}
Mark Everingham, Luc Van~Gool, Christopher~KI Williams, John Winn, and Andrew
  Zisserman.
\newblock The pascal visual object classes ({VOC}) challenge.
\newblock {\em International Journal of Computer Vision}, 88(2):303--338, 2010.

\bibitem{gao2019note}
Jiyang Gao, Jiang Wang, Shengyang Dai, Li-Jia Li, and Ram Nevatia.
\newblock Note-{RCNN}: Noise tolerant ensemble {RCNN} for semi-supervised
  object detection.
\newblock In {\em Proceedings of the IEEE International Conference on Computer
  Vision}, pages 9508--9517, 2019.

\bibitem{girshick2015fast}
Ross Girshick.
\newblock Fast {R-CNN}.
\newblock In {\em Proceedings of the IEEE International Conference on Computer
  Vision}, pages 1440--1448, 2015.

\bibitem{girshick2014rich}
Ross Girshick, Jeff Donahue, Trevor Darrell, and Jitendra Malik.
\newblock Rich feature hierarchies for accurate object detection and semantic
  segmentation.
\newblock In {\em Proceedings of the IEEE Conference on Computer Vision and
  Pattern Recognition}, pages 580--587, 2014.

\bibitem{He_2020_moco}
Kaiming He, Haoqi Fan, Yuxin Wu, Saining Xie, and Ross Girshick.
\newblock Momentum contrast for unsupervised visual representation learning.
\newblock In {\em Proceedings of the IEEE/CVF Conference on Computer Vision and
  Pattern Recognition}, June 2020.

\bibitem{hoffman2014lsda}
Judy Hoffman, Sergio Guadarrama, Eric~S Tzeng, Ronghang Hu, Jeff Donahue, Ross
  Girshick, Trevor Darrell, and Kate Saenko.
\newblock {LSDA}: Large scale detection through adaptation.
\newblock In {\em Advances in Neural Information Processing Systems}, pages
  3536--3544, 2014.

\bibitem{iscen2019label}
Ahmet Iscen, Giorgos Tolias, Yannis Avrithis, and Ondrej Chum.
\newblock Label propagation for deep semi-supervised learning.
\newblock In {\em Proceedings of the IEEE Conference on Computer Vision and
  Pattern Recognition}, pages 5070--5079, 2019.

\bibitem{izmailov2018averaging}
Pavel Izmailov, Dmitrii Podoprikhin, Timur Garipov, Dmitry Vetrov, and
  Andrew~Gordon Wilson.
\newblock Averaging weights leads to wider optima and better generalization.
\newblock {\em arXiv preprint arXiv:1803.05407}, 2018.

\bibitem{jeong2019consistency}
Jisoo Jeong, Seungeui Lee, Jeesoo Kim, and Nojun Kwak.
\newblock Consistency-based semi-supervised learning for object detection.
\newblock In {\em Advances in Neural Information Processing Systems}, pages
  10759--10768, 2019.

\bibitem{jeong2020interpolation}
Jisoo Jeong, Vikas Verma, Minsung Hyun, Juho Kannala, and Nojun Kwak.
\newblock Interpolation-based semi-supervised learning for object detection.
\newblock {\em arXiv preprint arXiv:2006.02158}, 2020.

\bibitem{laine2016temporal}
Samuli Laine and Timo Aila.
\newblock Temporal ensembling for semi-supervised learning.
\newblock {\em arXiv preprint arXiv:1610.02242}, 2016.

\bibitem{law2018cornernet}
Hei Law and Jia Deng.
\newblock {CornerNet}: Detecting objects as paired keypoints.
\newblock In {\em Proceedings of the European Conference on Computer Vision},
  pages 734--750, 2018.

\bibitem{lee2013pseudo}
Dong-Hyun Lee.
\newblock Pseudo-label: The simple and efficient semi-supervised learning
  method for deep neural networks.
\newblock In {\em Workshop on Challenges in Representation Learning, ICML},
  volume~3, 2013.

\bibitem{lin2017feature}
Tsung-Yi Lin, Piotr Doll{\'a}r, Ross Girshick, Kaiming He, Bharath Hariharan,
  and Serge Belongie.
\newblock Feature pyramid networks for object detection.
\newblock In {\em Proceedings of the IEEE Conference on Computer Vision and
  Pattern Recognition}, pages 2117--2125, 2017.

\bibitem{lin2014microsoft}
Tsung-Yi Lin, Michael Maire, Serge Belongie, James Hays, Pietro Perona, Deva
  Ramanan, Piotr Doll{\'a}r, and C~Lawrence Zitnick.
\newblock Microsoft {COCO}: Common objects in context.
\newblock In {\em Proceedings of the European Conference on Computer Vision},
  pages 740--755. Springer, 2014.

\bibitem{liu2016ssd}
Wei Liu, Dragomir Anguelov, Dumitru Erhan, Christian Szegedy, Scott Reed,
  Cheng-Yang Fu, and Alexander~C Berg.
\newblock {SSD}: Single shot multibox detector.
\newblock In {\em Proceedings of the European Conference on Computer Vision},
  pages 21--37. Springer, 2016.

\bibitem{mahajan2018exploring}
Dhruv Mahajan, Ross Girshick, Vignesh Ramanathan, Kaiming He, Manohar Paluri,
  Yixuan Li, Ashwin Bharambe, and Laurens van~der Maaten.
\newblock Exploring the limits of weakly supervised pretraining.
\newblock In {\em Proceedings of the European Conference on Computer Vision},
  pages 181--196, 2018.

\bibitem{newell2016stacked}
Alejandro Newell, Kaiyu Yang, and Jia Deng.
\newblock Stacked hourglass networks for human pose estimation.
\newblock In {\em Proceedings of the European Conference on Computer Vision},
  pages 483--499. Springer, 2016.

\bibitem{radosavovic2018data}
Ilija Radosavovic, Piotr Doll{\'a}r, Ross Girshick, Georgia Gkioxari, and
  Kaiming He.
\newblock Data distillation: Towards omni-supervised learning.
\newblock In {\em Proceedings of the IEEE Conference on Computer Vision and
  Pattern Recognition}, pages 4119--4128, 2018.

\bibitem{redmon2016you}
Joseph Redmon, Santosh Divvala, Ross Girshick, and Ali Farhadi.
\newblock You only look once: Unified, real-time object detection.
\newblock In {\em Proceedings of the IEEE Conference on Computer Vision and
  Pattern Recognition}, pages 779--788, 2016.

\bibitem{ren2015faster}
Shaoqing Ren, Kaiming He, Ross Girshick, and Jian Sun.
\newblock Faster {R-CNN}: Towards real-time object detection with region
  proposal networks.
\newblock In {\em Advances in Neural Information Processing Systems}, pages
  91--99, 2015.

\bibitem{riloff1996automatically}
Ellen Riloff.
\newblock Automatically generating extraction patterns from untagged text.
\newblock In {\em Proceedings of the National Conference on Artificial
  Intelligence}, pages 1044--1049, 1996.

\bibitem{rokach2010ensemble}
Lior Rokach.
\newblock Ensemble-based classifiers.
\newblock {\em Artificial Intelligence Review}, 33(1-2):1--39, 2010.

\bibitem{rosenberg2005semi}
Chuck Rosenberg, Martial Hebert, and Henry Schneiderman.
\newblock Semi-supervised self-training of object detection models.
\newblock In {\em 2005 Seventh IEEE Workshops on Applications of Computer
  Vision (WACV/MOTION'05) - Volume 1}, volume~1, pages 29--36, 2005.

\bibitem{russakovsky2015imagenet}
Olga Russakovsky, Jia Deng, Hao Su, Jonathan Krause, Sanjeev Satheesh, Sean Ma,
  Zhiheng Huang, Andrej Karpathy, Aditya Khosla, Michael Bernstein, et~al.
\newblock {ImageNet} large scale visual recognition challenge.
\newblock {\em International Journal of Computer Vision}, 115(3):211--252,
  2015.

\bibitem{sajjadi2016regularization}
Mehdi Sajjadi, Mehran Javanmardi, and Tolga Tasdizen.
\newblock Regularization with stochastic transformations and perturbations for
  deep semi-supervised learning.
\newblock {\em Advances in Neural Information Processing Systems},
  29:1163--1171, 2016.

\bibitem{scudder1965probability}
Henry Scudder.
\newblock Probability of error of some adaptive pattern-recognition machines.
\newblock {\em IEEE Transactions on Information Theory}, 11(3):363--371, 1965.

\bibitem{shorten2019survey}
Connor Shorten and Taghi~M Khoshgoftaar.
\newblock A survey on image data augmentation for deep learning.
\newblock {\em Journal of Big Data}, 6(1):1--48, 2019.

\bibitem{sohn2020fixmatch}
Kihyuk Sohn, David Berthelot, Nicholas Carlini, Zizhao Zhang, Han Zhang,
  Colin~A Raffel, Ekin~Dogus Cubuk, Alexey Kurakin, and Chun-Liang Li.
\newblock {FixMatch}: Simplifying semi-supervised learning with consistency and
  confidence.
\newblock In H. Larochelle, M. Ranzato, R. Hadsell, M.~F. Balcan, and H. Lin,
  editors, {\em Advances in Neural Information Processing Systems}, volume~33,
  pages 596--608. Curran Associates, Inc., 2020.

\bibitem{sohn2020simple}
Kihyuk Sohn, Zizhao Zhang, Chun-Liang Li, Han Zhang, Chen-Yu Lee, and Tomas
  Pfister.
\newblock A simple semi-supervised learning framework for object detection.
\newblock {\em arXiv preprint arXiv:2005.04757}, 2020.

\bibitem{sun2018integral}
Xiao Sun, Bin Xiao, Fangyin Wei, Shuang Liang, and Yichen Wei.
\newblock Integral human pose regression.
\newblock In {\em Proceedings of the European Conference on Computer Vision},
  pages 529--545, 2018.

\bibitem{tang2021proposal}
Peng Tang, Chetan Ramaiah, Yan Wang, Ran Xu, and Caiming Xiong.
\newblock Proposal learning for semi-supervised object detection.
\newblock In {\em Proceedings of the IEEE/CVF Winter Conference on Applications
  of Computer Vision}, pages 2291--2301, 2021.

\bibitem{tang2016large}
Yuxing Tang, Josiah Wang, Boyang Gao, Emmanuel Dellandr{\'e}a, Robert
  Gaizauskas, and Liming Chen.
\newblock Large scale semi-supervised object detection using visual and
  semantic knowledge transfer.
\newblock In {\em Proceedings of the IEEE Conference on Computer Vision and
  Pattern Recognition}, pages 2119--2128, 2016.

\bibitem{tarvainen2017mean}
Antti Tarvainen and Harri Valpola.
\newblock Mean teachers are better role models: Weight-averaged consistency
  targets improve semi-supervised deep learning results.
\newblock In {\em Advances in Neural Information Processing Systems}, pages
  1195--1204, 2017.

\bibitem{wang2018towards}
Keze Wang, Xiaopeng Yan, Dongyu Zhang, Lei Zhang, and Liang Lin.
\newblock Towards human-machine cooperation: Self-supervised sample mining for
  object detection.
\newblock In {\em Proceedings of the IEEE Conference on Computer Vision and
  Pattern Recognition}, pages 1605--1613, 2018.

\bibitem{xie2019unsupervised}
Qizhe Xie, Zihang Dai, Eduard Hovy, Minh-Thang Luong, and Quoc~V Le.
\newblock Unsupervised data augmentation for consistency training.
\newblock {\em arXiv preprint arXiv:1904.12848}, 2019.

\bibitem{Xie_2020_noisy_student}
Qizhe Xie, Minh-Thang Luong, Eduard Hovy, and Quoc~V. Le.
\newblock Self-training with noisy student improves {ImageNet} classification.
\newblock In {\em Proceedings of the IEEE/CVF Conference on Computer Vision and
  Pattern Recognition}, June 2020.

\bibitem{yarowsky1995unsupervised}
David Yarowsky.
\newblock Unsupervised word sense disambiguation rivaling supervised methods.
\newblock In {\em 33rd Annual Meeting of the Association for Computational
  Linguistics}, pages 189--196, 1995.

\bibitem{zhai2019s4l}
Xiaohua Zhai, Avital Oliver, Alexander Kolesnikov, and Lucas Beyer.
\newblock {S4L}: Self-supervised semi-supervised learning.
\newblock In {\em Proceedings of the IEEE International Conference on Computer
  Vision}, pages 1476--1485, 2019.

\bibitem{zhang2015lmdis}
Yuting Zhang, Kihyuk Sohn, Ruben Villegas, Gang Pan, and Honglak Lee.
\newblock Improving object detection with deep convolutional networks via
  {Bayesian} optimization and structured prediction.
\newblock In {\em {IEEE} Conference on Computer Vision and Pattern
  Recognition}, pages 249--258, June 2015.

\bibitem{zhu2002learning}
Xiaojin Zhu and Zoubin Ghahramani.
\newblock Learning from labeled and unlabeled data with label propagation.
\newblock {\em CMU CALD Tech Report CMU-CALD-02-107}, 2002.

\bibitem{zoph2020rethinking}
Barret Zoph, Golnaz Ghiasi, Tsung-Yi Lin, Yin Cui, Hanxiao Liu, Ekin~Dogus
  Cubuk, and Quoc Le.
\newblock Rethinking pre-training and self-training.
\newblock In H. Larochelle, M. Ranzato, R. Hadsell, M.~F. Balcan, and H. Lin,
  editors, {\em Advances in Neural Information Processing Systems}, volume~33,
  pages 3833--3845. Curran Associates, Inc., 2020.

\bibitem{zou2019learning}
Xu Zou, Sheng Zhong, Luxin Yan, Xiangyun Zhao, Jiahuan Zhou, and Ying Wu.
\newblock Learning robust facial landmark detection via hierarchical structured
  ensemble.
\newblock In {\em Proceedings of the IEEE International Conference on Computer
  Vision}, pages 141--150, 2019.

\end{thebibliography}


\begin{thebibliography}{1}\itemsep=-1pt

\bibitem{devries2017improved}
Terrance DeVries and Graham~W Taylor.
\newblock Improved regularization of convolutional neural networks with cutout.
\newblock {\em arXiv preprint arXiv:1708.04552}, 2017.

\bibitem{he2016deep}
Kaiming He, Xiangyu Zhang, Shaoqing Ren, and Jian Sun.
\newblock Deep residual learning for image recognition.
\newblock In {\em Proceedings of the IEEE Conference on Computer Vision and
  Pattern Recognition}, pages 770--778, 2016.

\bibitem{lin2017feature}
Tsung-Yi Lin, Piotr Doll{\'a}r, Ross Girshick, Kaiming He, Bharath Hariharan,
  and Serge Belongie.
\newblock Feature pyramid networks for object detection.
\newblock In {\em Proceedings of the IEEE Conference on Computer Vision and
  Pattern Recognition}, pages 2117--2125, 2017.

\bibitem{ren2015faster}
Shaoqing Ren, Kaiming He, Ross Girshick, and Jian Sun.
\newblock Faster {R-CNN}: Towards real-time object detection with region
  proposal networks.
\newblock In {\em Advances in Neural Information Processing Systems}, pages
  91--99, 2015.

\bibitem{sohn2020simple}
Kihyuk Sohn, Zizhao Zhang, Chun-Liang Li, Han Zhang, Chen-Yu Lee, and Tomas
  Pfister.
\newblock A simple semi-supervised learning framework for object detection.
\newblock {\em arXiv preprint arXiv:2005.04757}, 2020.

\end{thebibliography}
}
\end{document}


\title{Humble Teachers Teach Better Students for Semi Supervised Object Detection: Supplementary Materials}

\author{Yihe Tang \quad Weifeng Chen \quad Yijun Luo \quad Yuting Zhang\\
Amazon Web Services\\
{\tt\small tangacademic@gmail.com \quad \{weifec,yijunl,yutingzh\}@amazon.com}
}
\maketitle

\thispagestyle{empty}

\section{Weight of Unsupervised Loss}
\label{sec:ours_param}

This section studies how tuning the weight of the unsupervised loss changes the model performance and documents our decision for some low-level details.
Here we focus on the most important hyper-parameter defined in our model: the weight $\beta$ of the unsupervised loss $L_U$.

All the experiments in this section are conducted on \textit{MS-COCO train} with 10\% labeled.
We use the label-unlabeled data split 1 generated by code with the given random seed from \cite{sohn2020simple}.
We use Faster R-CNN~\cite{ren2015faster} with FPN~\cite{lin2017feature} and ResNet-50~\cite{he2016deep} as our base model.
It is worth noting that in this experiment, the number of region proposals used in the second stage of detection is 512, instead of 640 in the main paper and in Sec.~\ref{sec:loc}.

The final loss $L$ of our model is the sum of the supervised loss $L_{S}$ and the unsupervised loss $L_U$ with a scaling factor $\beta \frac{n_U}{n_S}$, as shown in Eqn.~\ref{eqn:beta}.
$n_U, n_S$ are the numbers of unlabeled and labeled images, and $\beta$ is an additional weight on unsupervised loss.
We use $\beta \frac{n_U}{n_S}$ to balance the unsupervised loss and the supervised loss in the model training.

\begin{align}
    L &= L_{S} + \beta \frac{n_U}{n_S} L_{U} \label{eqn:beta}
\end{align}

Tab.~\ref{tab:beta} reports the detailed ablation results.
We found that the performance deteriorates when $\beta$ is too small or too large, indicating that a balance between supervised learning and unsupervised learning is crucial to good performance for our Humble Teacher.
According to the ablation study, we set $\beta = 0.5$ across all experiments in the main paper and did not optimize them for particular experiments.

\begin{table}[htbp]
  \centering
  \resizebox{\columnwidth}{!}{
      \begin{tabular}{lcccccccc}
        \toprule
        $\beta$ & 0.1 & 0.2 & 0.3 & 0.4 & 0.5 & 0.6 & 0.7 & 0.8 \\
        \midrule
        mAP & 28.59 & 30.10 & 31.24 & 31.57 & 31.64 & 31.53 & 30.47 & 29.48  \\
        \bottomrule
      \end{tabular}
  }
  \caption{The results of models with different unsupervised weight $\beta$ on 10\% labeled \textit{MS-COCO 2017 train} (split 1), evaluated on the \textit{MS-COCO 2017 val} set.}
  \label{tab:beta}
\end{table}

\section{Ablation on Unsupervised Localization}
\label{sec:loc}

We study whether the unsupervised loss on bounding box regression heads improves the performance of the final model.
In this experiment, we compare enabling and disabling both bounding box regression heads for unsupervised loss.
We keep other parameters the same.
The basic experimental setup follows Sec.~\ref{sec:ours_param}, except the number of region proposals used in the second stage of detection is set to 640, following the main paper.
The results in Tab.~\ref{tab:loc} show that unsupervised learning on localization improves the final performance.

\begin{table}[htbp]
  \centering
  \begin{tabular}{lcc}
    \toprule
    Model & with localization & without localization\\
    \midrule
    mAP & 31.83 & 30.78 \\
    \bottomrule
  \end{tabular}
  \caption{Comparison between models with unsupervised localization enabled and disabled. The models are trained on 10\% labeled \textit{MS-COCO 2017 train} (split 1), evaluated on the \textit{MS-COCO 2017 val} set.}
  \label{tab:loc}
\end{table}

\section{Hard Label Experiments}
This section studies how different hyper-parameters impact the performance of models using hard labels.
As we compare our models with the hard label models in the main paper,
for fair comparison, we believe it is important to understand how hard label models perform the best.

There are two hyper-parameters for the hard label experiments: the weight $\beta$ of pseudo-label loss, and the confidence threshold $\theta$ as the threshold for accepting hard pseudo-labels as training samples.

\subsection{$\beta$: the Weight of Pseudo-Label Loss}
We study how the weight of unsupervised loss $\beta$ affects the model performance.
Denote $S_S$ as the labeled dataset and $S_U$ as the unlabeled dataset.
As shown in Eqn.~\ref{eqn:hard1}, the total loss $L$ of a hard label model is the sum of two losses: the loss on labeled images $S_S$ and the loss on pseudo-labeled images $S_U$.
$L_\mathrm{rcnn}$ is the sum of standard Faster R-CNN losses.
As in Sec.~\ref{sec:ours_param}, $n_U, n_S$ are numbers of the unlabeled images and labeled images, where $\beta$ is an additional weight on unsupervised loss.

\begin{align}
    L &= L_\mathrm{rcnn}(x)\vert_{x \in S_S} + \beta \frac{n_U}{n_S} L_\mathrm{rcnn}(y)\vert_{y \in S_U} \label{eqn:hard1}
\end{align}

For all the experiments, we perform experiments on \textit{MS-COCO train} with 10\% data labeled and use label-unlabeled data split 1 generated by code and random seed from \cite{sohn2020simple}.
We adopt Faster R-CNN~\cite{ren2015faster} with FPN~\cite{lin2017feature} and ResNet-50~\cite{he2016deep} as our base model.

\begin{figure}[htbp!]
  \centering
  \includegraphics[width=\columnwidth]{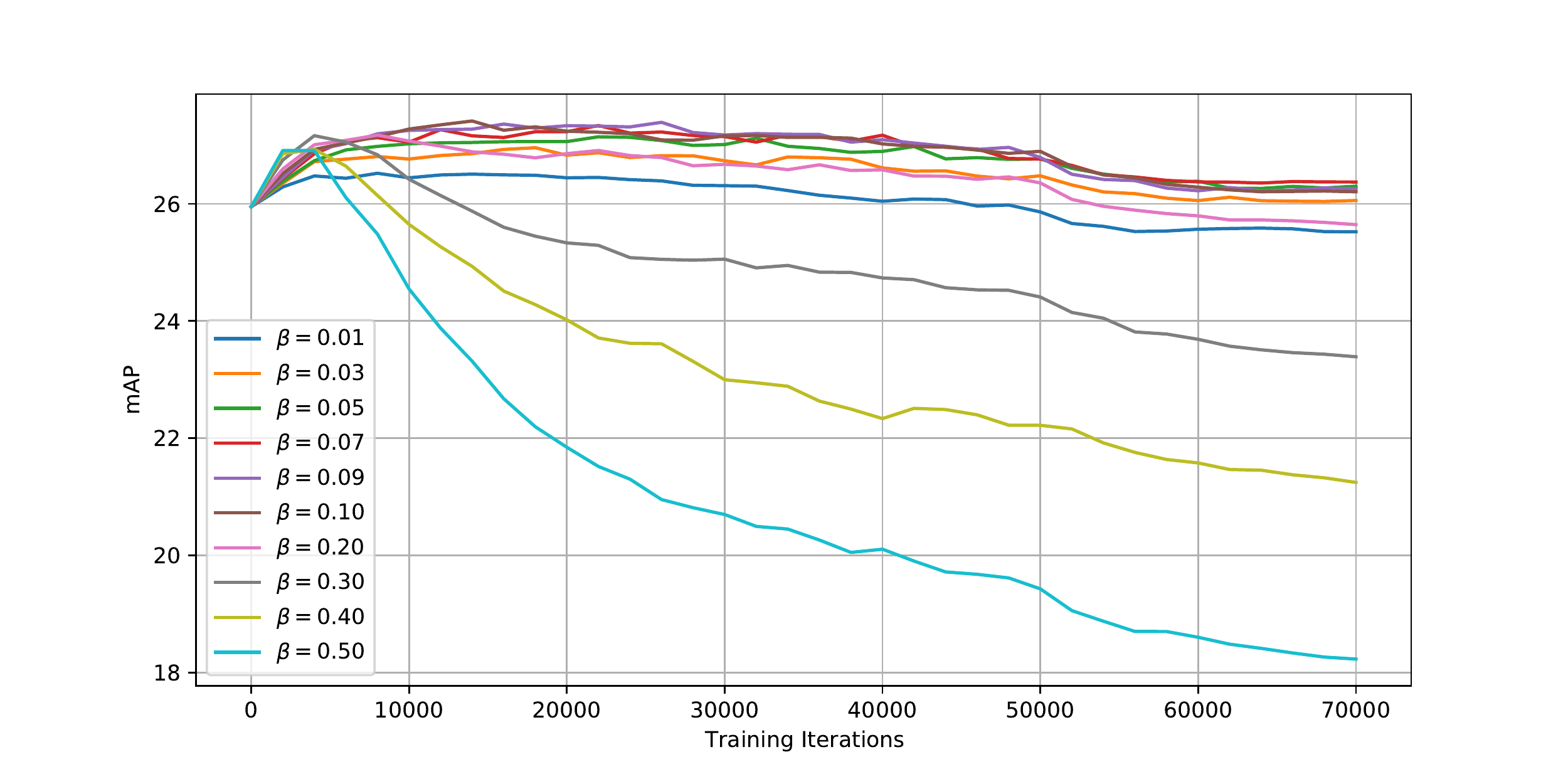}
  \caption{Comparison between hard label models with different $\beta$ trained on 10\% labeled \textit{MS-COCO 2017 train} (split 1), evaluated on the \textit{MS-COCO 2017 val} set.}
  \label{fig:hard_beta}
\end{figure}

The results are shown in Fig.~\ref{fig:hard_beta}.
We see that the model performs the best when $\beta$ is between 0.09 and 0.10.
For the overall best performance on different data splits, we used $\beta = 0.1$ for the hard label experiments in our main paper.
We find that the soft-label model outperforms the best hard-label model even with extensive parameter tuning for the hard label model directly on \emph{MS-COCO 2017 val}.

\subsection{$\theta$: the Confidence Threshold for Hard Labels}

In this section we study how the confidence threshold of hard labels affects the final model performance.
A reasonable confidence threshold for filtering low-quality pseudo-labels is crucial for hard label models.
Here we leave other parameters unchanged and only modify confidence threshold.

\begin{figure}[htbp!]
  \centering
  \includegraphics[width=\columnwidth]{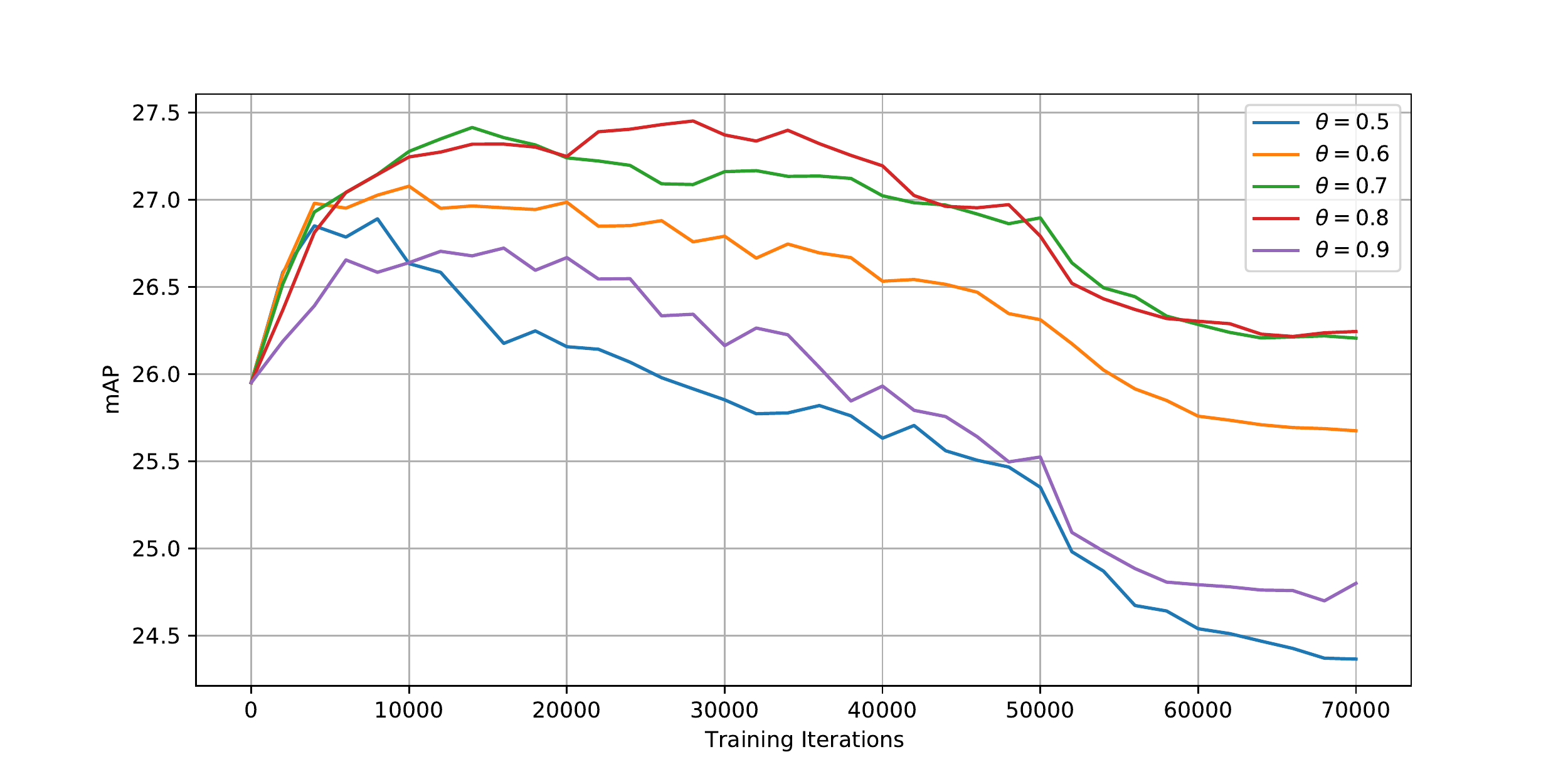}
  \caption{Comparison between hard label models with different $\theta$ trained on 10\% labeled \textit{MS-COCO 2017 train} (split 1), evaluated on the \textit{MS-COCO 2017 val} set.}
  \label{fig:hard_theta}
\end{figure}

The results in Fig.~\ref{fig:hard_theta} shows that the best threshold is between 0.7 and 0.8.
We select $\theta=0.7$ in the main paper as it leads to the best overall performance among five splits.
The result suggests that the soft-label model still outperforms the best hard-label model.

\section{Augmentation Details}

The strong augmentation we applied to our model follows~\cite{sohn2020simple}.
The augmentation consists of two operations: one operation changes the color, and another operation applies Cutout~\cite{devries2017improved}.
The configuration of the first operation is one randomly picked from the following operations, assuming all random numbers are sampled from uniform distributions:
\begin{enumerate}
    \item Identity: no changes at all.
    \item Apply Gaussian blurring with a standard deviation randomly taken from (0, 3).
    \item Apply average blurring by computing means over neighbourhoods. The kernel size is randomly picked from (2, 7).
    \item Sharpen a image and then alpha-blend the result with the original input image. The blending factor is randomly taken from (0, 1), where 0 means only the original image and 1 means only the sharpen image. The lightness/brightness of the sharpened image is taken from (0.75, 1.5).
    \item Apply noise sampled from Gaussian distributions elementwise to the input images. The means of the Gaussian distributions are set to 0.
    The standard deviations of the Gaussian distributions are sampled from (0, 0.05), which is relative to the maximum pixel value in the image format. Noise is applied on 50\% of images per-channel. 
    \item Invert the color with 5\% of probability.
    \item Add a value randomly taken from (-10, 10) to 50\% of image pixels per channel.
    \item Multiply each pixel with a value sampled from (0.5, 1.5). This operation applies to 50\% of pixels per channel.
    \item Multiply the contrast by a value randomly taken from (0.5, 2) per channel for the given input image.
\end{enumerate}

The second operation is a Cutout.
For each image, fill a random number $\alpha$ of cutout square patches on the original image with size either 0 or 0.2 of the input image height.
A Cutout patch with size 0 means that particular patch is canceled.
$\alpha$ is randomly taken from (1, 5).

{\small
\bibliographystyle{ieee_fullname}
\bibliography{egbib}
}